\documentclass{article} % For LaTeX2e
\usepackage[T1]{fontenc}
\usepackage{iclr2027_conference,times}

\usepackage{amsmath,amsfonts,bm}

\def\eqref#1{equation~\ref{#1}}
\def\1{\bm{1}}

\DeclareMathAlphabet{\mathsfit}{\encodingdefault}{\sfdefault}{m}{sl}
\SetMathAlphabet{\mathsfit}{bold}{\encodingdefault}{\sfdefault}{bx}{n}

\usepackage{hyperref}
\usepackage{url}
\usepackage{graphicx}
\usepackage{float}
\usepackage{booktabs}
\usepackage{amsmath,amssymb}
\usepackage{amsthm}
\newtheorem{proposition}{Proposition}[section]

\usepackage{microtype}
\usepackage{xcolor}
\usepackage{capt-of}
\newsavebox{\mechanismassetbox}

\newcommand{\mechanismcrop}[4]{%
  \begingroup
  \edef\mechanismimage{\noexpand\includegraphics[
    trim={\the\dimexpr#1\relax\space\the\dimexpr#2\relax\space\the\dimexpr#3\relax\space\the\dimexpr#4\relax},
    clip,width=\the\linewidth]{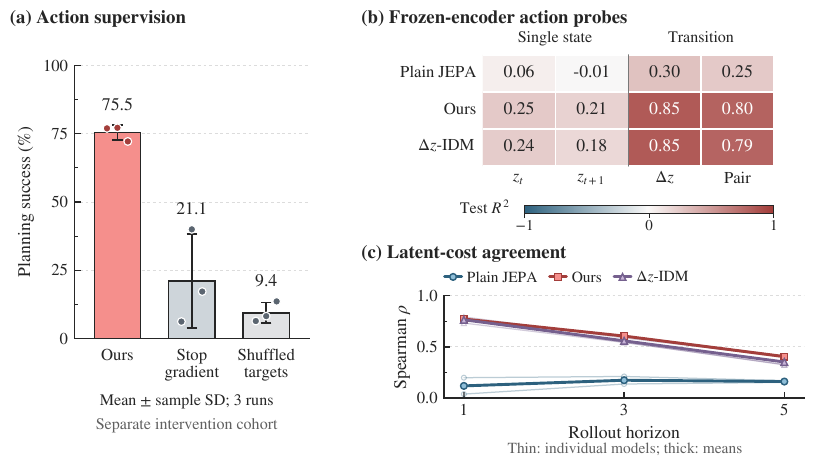}}%
  \mechanismimage
  \endgroup
}

\iclrfinalcopy

\title{Learning Intrinsic Dynamics in Latent World Models}

\author{\small Jiamin Shi$^1$ \quad Yuanhong Zhang$^1$ \quad Zhiyang Liu$^2$ \quad Jiawei Sun$^2$ \quad Jingwen Fu$^3$\\[3pt]
\normalfont\small $^1$Xi'an Jiaotong University \quad $^2$National University of Singapore\\
\normalfont\small $^3$Beijing Zhongguancun Academy}

\newcommand{\PTPlainSuccess}{$9.1{\pm}2.7$}
\newcommand{\PTPlainNFE}{$1.353{\pm}0.108$}
\newcommand{\PTDeltaSuccess}{$65.8{\pm}14.7$}
\newcommand{\PTDeltaNFE}{$0.777{\pm}0.177$}
\newcommand{\PTConcatSuccess}{$77.2{\pm}0.2$}
\newcommand{\PTConcatNFE}{$0.652{\pm}0.032$}
\newcommand{\PTLowPlainSuccess}{$72.2{\pm}7.9$}
\newcommand{\PTLowConcatSuccess}{$87.5{\pm}4.4$}

\newcommand{\PTAssignConcatRange}{$74.2$--$78.4$}
\newcommand{\PTAssignPlainRange}{$5.6$--$61.2$}

\newcommand{\PTSuccessInteraction}{$+52.9{\pm}7.1$}

\newcommand{\TRLowPlainSuccess}{$88.5{\pm}1.3$}
\newcommand{\TRLowConcatSuccess}{$91.1{\pm}1.0$}
\newcommand{\CubeLowPlainSuccess}{$60.9{\pm}1.2$}
\newcommand{\CubeLowConcatSuccess}{$71.1{\pm}2.2$}
\newcommand{\ReacherLowPlainSuccess}{$100.0{\pm}0.0$}
\newcommand{\ReacherLowConcatSuccess}{$100.0{\pm}0.0$}
\newcommand{\TRPlainSuccess}{$53.2{\pm}1.4$}
\newcommand{\TRPlainNFE}{$0.798{\pm}0.015$}
\newcommand{\TRDeltaSuccess}{$97.3{\pm}2.4$}
\newcommand{\TRConcatSuccess}{$97.3{\pm}3.2$}
\newcommand{\TRConcatNFE}{$0.457{\pm}0.019$}

\newcommand{\PTPlainProgress}{$0.481{\pm}0.046$}
\newcommand{\PTConcatProgress}{$0.674{\pm}0.007$}
\newcommand{\TRPlainProgress}{$0.387{\pm}0.008$}
\newcommand{\TRConcatProgress}{$0.568{\pm}0.010$}

\newcommand{\PTPlainProbePair}{$0.251{\pm}0.209$}
\newcommand{\PTDeltaProbeDelta}{$0.850{\pm}0.002$}
\newcommand{\PTConcatProbeDelta}{$0.846{\pm}0.002$}
\newcommand{\PTConcatProbePair}{$0.799{\pm}0.001$}
\newcommand{\PTConcatSingleProbe}{$0.254{\pm}0.003$}

\begin{document}

\maketitle
\fancyhead{}
\renewcommand{\headrulewidth}{0pt}
\raggedbottom
\hypersetup{hidelinks}

\begin{abstract}
Learning intrinsic dynamics---the relationship between actions and their consequences---is essential for world models. However, visual variations unrelated to the underlying control process can obscure this relationship in learned representations.
We investigate the connection between \textit{action predictability}, defined as the ability to reconstruct an action from consecutive latent states, and the preservation of intrinsic dynamics.
Our central hypothesis is that \textit{encouraging action predictability discourages representations from discarding the distinctions between different actions and their consequences}.
We provide two complementary forms of evidence for this connection.
Theoretically, we derive a positive lower bound on squared inverse prediction error when the encoder merges distinguishable outcomes of different actions from the same state and appearance, and show that forward prediction can still be exact on those transitions.
Empirically, controlled appearance-diversity comparisons, supervision interventions, and latent probes support the role of action prediction in preserving control-relevant transition information.
We instantiate action predictability with Concat IDM, which predicts actions from concatenated latent states and updates the shared encoder using action labels already available in offline trajectories, without changing the test-time planner.
Across three independent training runs on Push-T with 24 appearances, Concat IDM increases planning success from $9.1\%$ to $77.2\%$; its paired improvement is $52.9$ percentage points larger than that observed with a single appearance.
Together, these findings support action predictability as an inductive bias toward intrinsic dynamics and demonstrate the practical benefits of Concat IDM for latent planning under visual diversity.

\end{abstract}

\section{Introduction}
\label{sec:intro}

\begin{figure}[H]
\centering
\begin{minipage}[t]{0.48\linewidth}
\centering
\vspace{0pt}
\includegraphics[width=\linewidth]{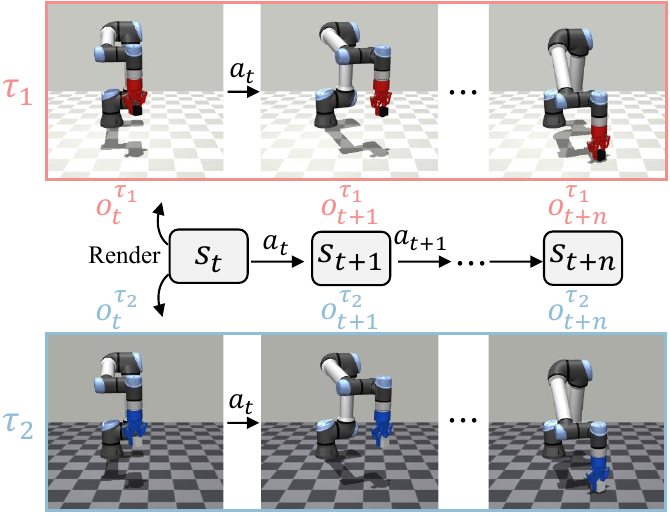}
\par\small (a) Same dynamics, different appearances
\end{minipage}\hfill
\begin{minipage}[t]{0.49\linewidth}
\centering
\vspace{0pt}
\includegraphics[width=\linewidth]{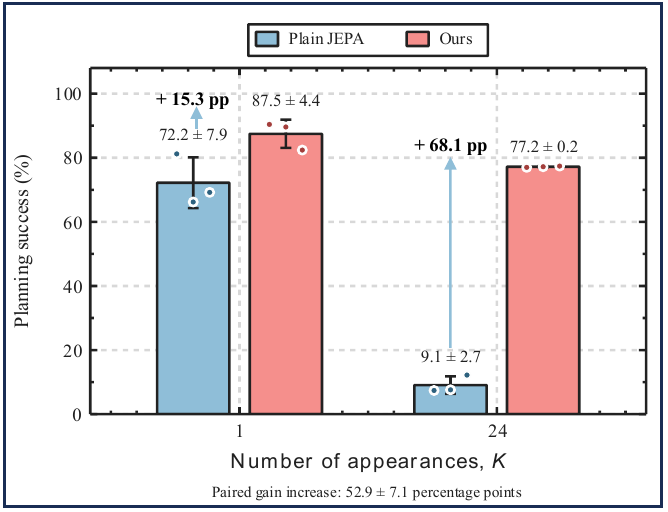}
\par\small (b) Larger gains under visual diversity
\end{minipage}
\caption{\textbf{Visual diversity amplifies the benefit of action prediction.}
(a) Schematic of appearance variation with physical trajectories held fixed.
(b) Under this controlled variation on Push-T, the gain of Concat IDM over Plain JEPA grows from $15.3$ pp at $K{=}1$ to $68.1$ pp at $K{=}24$.}
\label{fig:motivation}
\end{figure}
An action-conditioned latent world model should capture how the underlying system responds to actions.
We refer to this action-conditioned transition structure as \textbf{intrinsic dynamics}.
This relationship concerns the underlying control process rather than the visual appearance through which it is observed.
However, images contain variations unrelated to that process: object color, background, lighting, and rendering style can change observations without changing the physical transition.
We refer to these variations as \textbf{external factors} (Figure~\ref{fig:motivation}a).
Learning from visually diverse observations therefore presents a challenge: the representation must retain action-dependent transition information even as the appearance of the same underlying process varies.

When the encoder and predictor are trained jointly, the encoder also defines the prediction targets.
Accurate latent prediction does not, by itself, ensure that the representation preserves distinctions in how the underlying system responds to different actions.
We therefore investigate whether \textbf{action predictability} can help preserve these distinctions.
We define action predictability as the ability to reconstruct the executed action $a_t$ from consecutive latent states $z_t$ and $z_{t+1}$.
Predicting actions from transitions is an established inverse-dynamics objective in representation learning and latent planning~\citep{agrawal2016learning,pathak2017curiosity,paster2021planning,sobal2025pldm}.
Our focus is on understanding how this requirement can help preserve the relationship between actions and their consequences when external factors vary.

Our central hypothesis is that \emph{encouraging action predictability helps representations distinguish the consequences of different actions}.
Suppose two different actions taken from the same state lead to different outcomes.
If the encoder maps both outcomes to the same latent state, the latent transition no longer tells us which action was taken.
Training the model to predict the action therefore encourages the encoder to keep these outcomes distinct.

We provide two complementary forms of evidence connecting action predictability to learning intrinsic dynamics. \textbf{Theoretically}, if the encoder maps distinguishable outcomes of different actions from the same state and appearance to the same latent state, squared inverse prediction error has a positive lower bound, while forward prediction can remain exact (Proposition~\ref{prop:merged_transitions}). Action prediction thus penalizes the loss of action--consequence distinctions that forward prediction may overlook. \textbf{Empirically}, controlled comparisons show larger planning gains from action prediction under greater appearance diversity, with physical trajectories held fixed. Supervision interventions link these gains to encoder updates from correctly paired action labels, while frozen-encoder probes show that actions are decoded more accurately from transitions than from either endpoint alone. Together, these findings support action predictability as an inductive bias toward intrinsic dynamics, rather than a guarantee of complete dynamics recovery.

We instantiate action predictability with \textbf{Concat IDM}, which predicts $a_t$ from concatenated latent states $[z_t;z_{t+1}]$.
Its inverse-dynamics loss updates the shared encoder, so that action supervision directly constrains the representation used by the world model.
The method uses action labels already available in offline trajectories and leaves the test-time planner unchanged.
We implement Concat IDM on the LeWorldModel backbone~\citep{maes2026leworldmodel} and compare it against Plain JEPA, which uses the same backbone without the inverse-dynamics objective. Across three independent training runs on Push-T with 24 appearances, Concat IDM increases planning success from $9.1\%$ to $77.2\%$.
Its paired improvement over Plain JEPA is $52.9$ percentage points larger than that observed with a single appearance (Figure~\ref{fig:motivation}b).
Because the physical trajectory pool is held fixed, this comparison shows that the benefit grows sharply with visual diversity in this controlled setting.
These results demonstrate the practical value of Concat IDM while complementing the theoretical and empirical evidence for action predictability as an inductive bias toward intrinsic dynamics.

The \textbf{central contribution} of this paper is to clarify the connection between action predictability and intrinsic dynamics in latent world models. Specifically, we \textbf{(1)} hypothesize that encouraging action predictability helps preserve distinctions between the consequences of different actions that action-conditioned forward prediction alone may overlook; \textbf{(2)} provide theoretical and empirical evidence supporting this hypothesis; and \textbf{(3)} instantiate this principle with Concat IDM and evaluate its effects on learned representations and downstream planning.

\section{Related Work}
\label{sec:related}

\paragraph{Learning latent dynamics.}
Latent world models support online planning in PlaNet, behavior learning through imagination in Dreamer, and task-oriented predictive control in TD-MPC~\citep{hafner2019planet,hafner2020dreamer,hansen2022tdmpc}.
Self-predictive representations also use multi-step latent prediction as an auxiliary objective for visual RL~\citep{schwarzer2021spr}.
Joint-embedding predictive architectures learn prediction targets in representation space without pixel reconstruction~\citep{assran2023self,bardes2024vjepa}; V-JEPA~2 extends video pretraining to action-conditioned robotic planning~\citep{assran2025vjepa2}.
DINO-WM predicts frozen DINOv2 features~\citep{zhou2025dinowm}, whereas PLDM and LeWorldModel jointly learn the encoder and latent dynamics~\citep{sobal2025pldm,maes2026leworldmodel}.
PLDM already combines inverse dynamics with VICReg~\citep{bardes2022vicreg}; LeWorldModel combines next-embedding prediction with SIGReg~\citep{balestriero2025lejepa}.
Concurrent Temporal-Distance JEPA addresses the mismatch between short-horizon representation learning and multi-step planning costs~\citep{bai2026tdjepa}.
We study a different axis: how visual diversity over fixed physical trajectories affects jointly learned representations and the benefit of inverse supervision.

\paragraph{Inverse dynamics and action predictability.}
Inverse dynamics provides supervision for representations of physical interaction and control~\citep{agrawal2016learning,pathak2017curiosity,paster2021planning}.
Multistep inverse models underlie provable RL with exogenous distractors~\citep{efroni2022exogenous}, control-endogenous state discovery in AC-State~\citep{lamb2023acstate}, and offline agent-centric representation learning in ACRO~\citep{islam2023acro}.
ACDF identifies failures of multistep inverse prediction alone and combines it with forward modeling~\citep{levine2024multistep}.
Our analysis isolates how inverse prediction penalizes the merging of distinguishable action consequences.
Concurrent Delta-JEPA predicts actions from latent displacements~\citep{zhang2026deltajepa}; our matched $\Delta z$-IDM comparison changes only the inverse input, holding the backbone, regularizer, data, and planner fixed.
Concurrent ACID uses inverse dynamics at decision time to score imagined transitions~\citep{seo2026acid}.
In contrast, our inverse loss constrains the shared encoder during training; the inverse head is discarded for planning.

\paragraph{External visual factors.}
Domain randomization and adversarial domain learning address visual variation~\citep{tobin2017domain,ganin2016domain}.
CURL learns contrastive representations from augmented views~\citep{laskin2020curl}, while RAD and DrQ improve pixel-based control through data augmentation~\citep{laskin2020rad,kostrikov2020drq}.
Bisimulation-based approaches preserve control-relevant state equivalence~\citep{zhang2021learning}; \citet{toso2026invariant} apply this principle over pretrained visual features for latent world-model planning under background changes and distractors.
Concurrent ACPC diagnoses predictive consistency between matched clean and perturbed histories under the same actions~\citep{an2026acpc}.
We study action--consequence distinctions during joint encoder--predictor training by varying visual appearances over fixed offline trajectories, without an explicit appearance-invariance objective or matched-view supervision.

\section{Theory and Method}
\label{sec:method}

\subsection{Intrinsic dynamics and external factors}
\label{sec:setting}

Let $s_t$ denote the physical state, $a_t$ the executed action, and $c$ the factors that affect only the visual observation.
For the controlled environments considered here,
\begin{equation}
o_t=R(s_t,c),
\qquad
s_{t+1}=F(s_t,a_t).
\label{eq:setting}
\end{equation}
The action-conditioned transition structure induced by~$F$ is the \textbf{intrinsic dynamics}; the variables in~$c$ are \textbf{external factors} because they alter the rendering without altering the physical transition.

We construct appearance-diverse datasets by rendering a fixed pool of control trajectories with a finite palette of size~$K$.
Each trajectory receives one rendering, and matched renderings of the same trajectory are not provided during training.
The physical state and action sequences are unchanged as~$K$ varies; increasing~$K$ redistributes the same control data across more visual realizations.
We also vary the episode-level assignment while holding the physical trajectory pool and palette fixed (Section~\ref{sec:assignment}).

\begin{figure}[t]
\centering
\begin{minipage}[t]{0.64\linewidth}
\centering
\vspace{0pt}
\includegraphics[width=\linewidth]{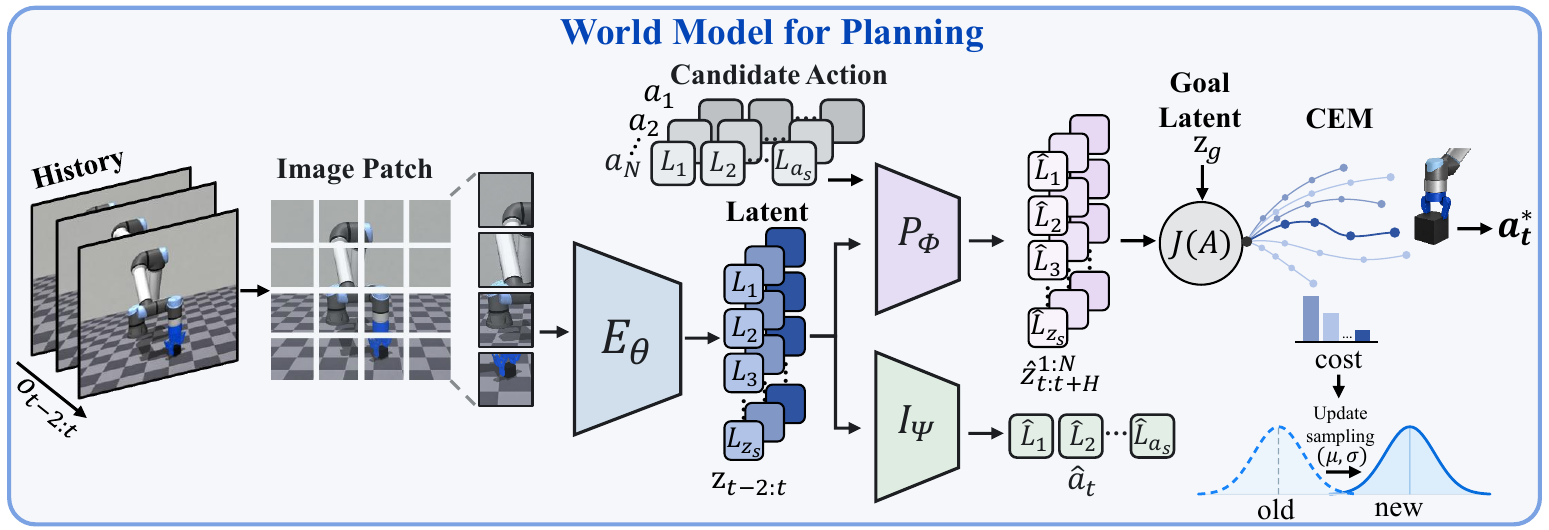}
\par\small (a) World model and planning
\end{minipage}\hfill
\begin{minipage}[t]{0.36\linewidth}
\centering
\vspace{0pt}
\includegraphics[width=\linewidth]{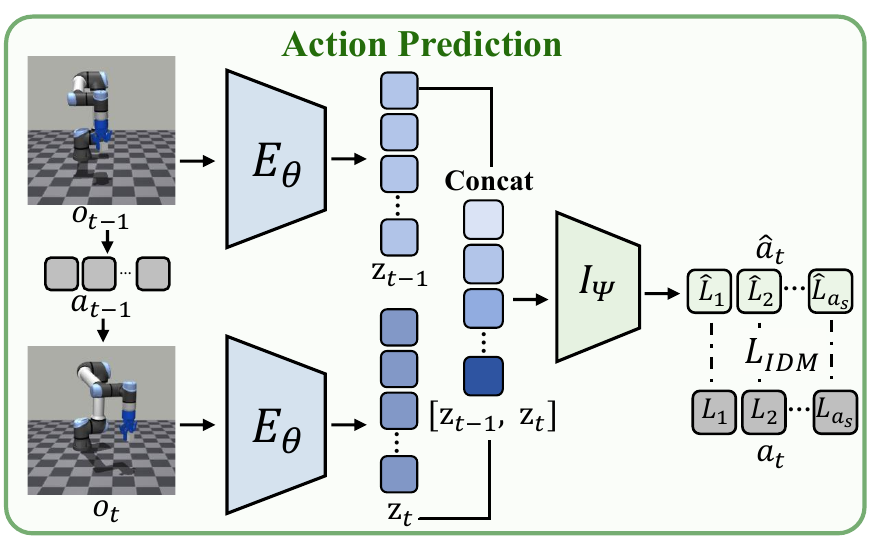}
\par\small (b) Action supervision
\end{minipage}
\caption{\textbf{Action-supervised latent world modeling and planning.}
$E_\theta$: visual encoder; $P_\phi$: latent predictor; $I_\psi$: inverse dynamics head.
(a) CEM ranks predicted rollouts by terminal latent goal cost.
(b) Concat IDM predicts actions from concatenated observed latents, supervising the shared encoder during training.}
\label{fig:overview}
\label{fig:setting}
\label{fig:method}
\end{figure}

\subsection{Action predictability and transition distinctions}
\label{sec:idm}

Action predictability constrains the information retained between consecutive states: their encodings should remain predictive of the executed action.
To characterize this constraint, consider a fixed encoder $E_\theta$ and an inverse decoder $g$ that predicts actions in $\mathbb{R}^{d_a}$.
Fix a physical state $s$ and appearance $c$ in Eq.~(\ref{eq:setting}).
Let $o=R(s,c)$ and $o_i^+=R(F(s,a_i),c)$ for distinct actions $a_1,a_2$.
Assume $o_1^+\ne o_2^+$, so the two outcomes are distinguishable before encoding.
Consider a distribution assigning probabilities $p\in(0,1)$ and $1-p$ to these two transitions, and define
\begin{equation}
u_i=[E_\theta(o);E_\theta(o_i^+)],\qquad
\mathcal{R}_{\mathrm{pair}}
=\frac{p\|g(u_1)-a_1\|_2^2+(1-p)\|g(u_2)-a_2\|_2^2}{d_a}.
\label{eq:pair_risk}
\end{equation}

\begin{proposition}[Penalty for merging distinguishable outcomes]
\label{prop:merged_transitions}
Under the conditions above, if $E_\theta(o_1^+)=E_\theta(o_2^+)$, then every inverse decoder satisfies
\begin{equation}
\mathcal{R}_{\mathrm{pair}}
\geq \frac{p(1-p)}{d_a}\|a_1-a_2\|_2^2>0.
\label{eq:merged_bound}
\end{equation}
The bound is attained by predicting $\bar a=pa_1+(1-p)a_2$ at the shared latent input.
For this pair alone, a forward predictor can nevertheless attain zero latent prediction error by outputting the common successor encoding for both actions.
\end{proposition}

Proposition~\ref{prop:merged_transitions} isolates an information-loss mode of latent prediction: distinct physical outcomes can share an exactly predictable latent target, whereas inverse supervision penalizes this merger.
The result holds at a fixed appearance; Section~\ref{sec:diversity} tests empirically how the benefit of action supervision changes as appearance diversity varies.
To see the bound, let $b$ be the decoder output at the shared latent input and write $\bar a=pa_1+(1-p)a_2$. Completing the square gives $p\lVert b-a_1\rVert_2^2+(1-p)\lVert b-a_2\rVert_2^2=\lVert b-\bar a\rVert_2^2+p(1-p)\lVert a_1-a_2\rVert_2^2$.

\paragraph{What the inverse objective preserves.}
The pairwise result extends to an arbitrary transition distribution with finite second action moments.
Write $X=(O_t,O_{t+1})$, $U=[E_\theta(O_t);E_\theta(O_{t+1})]$, and let $\mu_X=\mathbb{E}[A\mid X]$ and $\mu_U=\mathbb{E}[A\mid U]$.
If $\mathcal{R}^{*}_X$ and $\mathcal{R}^{*}_U$ are the minimum coordinate-averaged squared action errors over unrestricted decoders of $X$ and $U$, respectively, then
\begin{equation}
\mathcal{R}^{*}_U-\mathcal{R}^{*}_X
=\frac{1}{d_a}\mathbb{E}\big[\|\mu_X-\mu_U\|_2^2\big].
\label{eq:inverse_information_gap}
\end{equation}
Thus, encoding incurs an additional inverse-prediction penalty whenever it removes distinctions needed to recover the conditional action mean.
The decomposition separates ambiguity already present in the observations from information lost through the encoder.
Forward prediction rewards predictable latent targets, whereas inverse prediction makes those targets retain information about the actions that produced them.
This characterization motivates examining both action decoding and forward rollout consistency: the two diagnostics test different properties of the representation used by the planner.

\subsection{Concat IDM and training objective}

We use an action-conditioned JEPA with a shared visual encoder and a latent predictor.
Figure~\ref{fig:method} summarizes the planning pipeline and action-supervised training.
Our implementation follows LeWorldModel~\citep{maes2026leworldmodel}.
The encoder produces $z_t=E_\theta(o_t)$, and the predictor forecasts
\begin{equation}
\hat z_{t+1}=P_\phi(z_{\leq t},a_{\leq t}),
\end{equation}
where the observation history is limited to three frames.
Here and in the training losses below, $a_t$ concatenates the primitive actions between consecutive sampled observations, standardized coordinatewise using the offline dataset mean and standard deviation.
The next-embedding prediction loss is
\begin{equation}
\mathcal{L}_{\mathrm{pred}}
=
\frac{1}{N_{\mathrm{pred}}d_z}\sum_t
\left\|\hat z_{t+1}-z_{t+1}\right\|_2^2,
\end{equation}
where $N_{\mathrm{pred}}$ counts the predicted transitions and $d_z$ is the embedding dimension.
As in LeWorldModel, the same encoder produces context and target embeddings, and gradients pass through both branches; there is no target network or stop-gradient in this loss.
We use SIGReg~\citep{balestriero2025lejepa} to discourage representational collapse by regularizing random one-dimensional projections of the encoded latents toward a standard Gaussian distribution.
We denote this regularizer by $\mathcal{L}_{\mathrm{SIG}}$ and use $17$ knots and $1024$ projections.

We instantiate action predictability with an inverse dynamics head~$I_\psi$.
Concatenation makes both endpoint representations available to the decoder without restricting it to their displacement.
Our main variant, \textbf{Concat IDM}, predicts
\begin{equation}
\hat a_t=I_\psi\big([z_t;z_{t+1}]\big),
\qquad
\mathcal{L}_{\mathrm{idm}}
=
\frac{1}{N_{\mathrm{idm}}d_a}\sum_t
\left\|\hat a_t-a_t\right\|_2^2,
\label{eq:idm_loss}
\end{equation}
where $N_{\mathrm{idm}}$ counts adjacent latent pairs and $d_a$ is the action-target dimension.
Both squared losses are averaged over latent pairs and target coordinates.
The complete objective is
\begin{equation}
\mathcal{L}
=
\mathcal{L}_{\mathrm{pred}}
+
\beta\,\mathcal{L}_{\mathrm{SIG}}
+
\lambda\,\mathcal{L}_{\mathrm{idm}},
\label{eq:objective}
\end{equation}
with SIGReg weight $\beta{=}0.09$ and IDM weight $\lambda{=}1$ in the main experiments.
SIGReg regularizes the marginal latent distribution, while the inverse loss supervises action-dependent relations between endpoint encodings.
The inverse loss updates the IDM head and propagates through both endpoint encodings to the shared encoder; it does not update the forward predictor directly.

\paragraph{$\Delta z$-IDM.}
To compare input parameterizations, we replace the concatenated endpoints by their displacement and train $I_\psi(z_{t+1}-z_t)$ while keeping the remaining model and objective fixed.

\paragraph{Intervention ablations.}
(i)~the stop-gradient intervention replaces both IDM inputs with detached latents, allowing the inverse head to train while preventing its objective from updating the encoder.
(ii)~the shuffled-target intervention permutes action labels within each batch, preserving the architecture while breaking their correspondence with the transitions.
Together, these interventions test whether the gain depends on representation updates induced by correctly paired action supervision.

\subsection{Latent planning}

At test time, the encoder and predictor are frozen.
Given a goal observation~$o_g$, we compute $z_g=E_\theta(o_g)$.
Starting from the encoded observation history, the predictor rolls each candidate action sequence $A=(a_t,\ldots,a_{t+H-1})$ forward autoregressively to obtain~$\hat z_{t+H}$.
The main planner evaluates the terminal prediction in the full latent space:
\begin{equation}
J(A)=\left\|\hat z_{t+H}-z_g\right\|_2^2.
\label{eq:planning_cost}
\end{equation}
CEM iteratively updates a distribution over action sequences to minimize~$J$.
The agent executes the first action block from the selected sequence, receives a new observation, and replans in closed loop.
CEM uses $300$ samples, $30$ optimization steps, $30$ elites, planning and receding horizons of $5$, and an action block of $5$ primitive steps.

\section{Experiments}
\label{sec:exp}

\subsection{Experimental setup and metrics}

Our experiments ask whether action supervision improves latent planning, how its benefit depends on visual diversity, and what action information the learned representations retain.

\paragraph{Protocol.}
Primary planning comparisons use three independently trained models per condition, each evaluated on 500 episodes in 10 runs of 50 episodes.
Paired methods share the physical trajectory pool, evaluation tasks, and planner.
The Push-T diversity comparison pairs training runs across $K{=}1$ and $K{=}24$; the other tasks compare three-model means at $K{=}1$ and their task-specific high-diversity settings.
Evaluation starts and goals are sampled from the offline trajectory pool used for world-model learning, with current and goal observations sharing an appearance from the training palette.
Separate evaluations change the training appearance assignment or use a disjoint test palette.
The backbone uses a ViT-Tiny encoder with $192$-dimensional embeddings and a six-layer predictor; the IDM head has hidden width $512$. Unless noted, training uses AdamW for $10$ epochs, learning rate $2\times10^{-5}$, weight decay $10^{-3}$, and batch size $128$.

\paragraph{Metrics.}
Our primary metric is closed-loop success. Unless specified in the table, aggregate results are mean~$\pm$~sample SD across three independently trained models per condition or appearance assignment.
Action-decoding $R^2$ measures accessible action information; predicted-versus-realized latent-cost agreement measures rollout-ranking consistency.
Normalized final error is $\operatorname{NFE}=d(s_T,g)/(d(s_0,g)+10^{-8})$, where $s_T$ is the recorded state at first success or the end of the evaluation budget. Best progress is $1-\min_{0\leq t\leq T}d(s_t,g)/(d(s_0,g)+10^{-8})$. Push-T uses the Euclidean distance of the raw state vector (agent and block positions, block orientation, and agent velocity); TwoRoom uses agent-to-target Euclidean distance. These auxiliary metrics are compared within each task.

\subsection{Performance across control tasks}
\label{sec:main_results}

\begin{table}[t]
\caption{\textbf{Planning gains across appearance conditions.} Gain is Concat IDM minus Plain JEPA, computed from the displayed success means in percentage points (pp). Bold marks the higher success mean, including ties, and positive gains.}
\label{tab:main_multirun}
\centering
\small
\setlength{\tabcolsep}{3pt}
\begin{tabular*}{\linewidth}{@{\extracolsep{\fill}}lrrrcrrr@{}}
\toprule
& \multicolumn{3}{c}{Single appearance ($K{=}1$)} & \multicolumn{4}{c}{Multiple appearances} \\
\cmidrule(lr){2-4}\cmidrule(lr){5-8}
Task & Plain JEPA & Concat IDM & Gain & $K$ & Plain JEPA & Concat IDM & Gain \\
\midrule
Push-T & \PTLowPlainSuccess & \textbf{\boldmath\PTLowConcatSuccess} & $\mathbf{+15.3}$ & 24 & \PTPlainSuccess & \textbf{\boldmath\PTConcatSuccess} & $\mathbf{+68.1}$ \\
TwoRoom & \TRLowPlainSuccess & \textbf{\boldmath\TRLowConcatSuccess} & $\mathbf{+2.6}$ & 36 & \TRPlainSuccess & \textbf{\boldmath\TRConcatSuccess} & $\mathbf{+44.1}$ \\
Cube & \CubeLowPlainSuccess & \textbf{\boldmath\CubeLowConcatSuccess} & $\mathbf{+10.2}$ & 24 & $60.7{\pm}1.2$ & $\mathbf{75.4{\pm}1.3}$ & $\mathbf{+14.7}$ \\
Reacher & \textbf{\boldmath\ReacherLowPlainSuccess} & \textbf{\boldmath\ReacherLowConcatSuccess} & $0.0$ & 24 & $\mathbf{100.0{\pm}0.0}$ & $\mathbf{100.0{\pm}0.0}$ & $0.0$ \\
\bottomrule
\end{tabular*}
\end{table}

\paragraph{Planning gains across tasks.}
Concat IDM improves planning on Push-T, TwoRoom, and Cube without changing the test-time planner (Table~\ref{tab:main_multirun}).
Its advantage is larger under high visual diversity on Push-T and TwoRoom: gains rise from $15.3$ to $68.1$ percentage points and from $2.6$ to $44.1$ points, respectively.
On Cube, Plain JEPA remains near $61\%$ while Concat IDM gains $10.2$ points at $K{=}1$ and $14.7$ at $K{=}24$.
Reacher is saturated at $100\%$ for both methods.

The continuous-error results complement these success gains.
On Push-T at $K{=}24$, NFE falls from \PTPlainNFE\ to \PTConcatNFE, while best progress rises from \PTPlainProgress\ to \PTConcatProgress.
TwoRoom shows the same direction: NFE falls from \TRPlainNFE\ to \TRConcatNFE\ and best progress rises from \TRPlainProgress\ to \TRConcatProgress.
These within-task comparisons indicate improvement in both the closest approach to the goal and the final recorded state, rather than only a change in how often a success threshold is crossed.

\begin{figure}[t]
\centering
\begin{minipage}[t]{0.62\linewidth}
\vspace{0pt}
\captionof{table}{Planning success (\%) and normalized final error (NFE) for inverse-dynamics designs and regularizers on Push-T ($K{=}24$).}
\label{tab:main_baselines}
\label{tab:regularizer_2x2}
\centering
\small
\renewcommand{\arraystretch}{1.5}
\textbf{(a) IDM-based alternatives}\par\smallskip
\setlength{\tabcolsep}{3pt}
\begin{tabular*}{\linewidth}{@{\extracolsep{\fill}}lrr@{}}
\toprule
Method & Success (\%) $\uparrow$ & NFE $\downarrow$ \\
\midrule
Plain JEPA & \PTPlainSuccess & \PTPlainNFE \\
PLDM & $53.2{\pm}11.8$ & $0.783{\pm}0.103$ \\
$\Delta z$-IDM & \PTDeltaSuccess & \PTDeltaNFE \\
Concat IDM & \textbf{\boldmath\PTConcatSuccess} & \PTConcatNFE \\
\bottomrule
\end{tabular*}
\par\medskip
\textbf{(b) Regularizer control}\par\smallskip
\footnotesize
\setlength{\tabcolsep}{1.5pt}
\begin{tabular*}{\linewidth}{@{\extracolsep{\fill}}lrrrrr@{}}
\toprule
Reg. & \multicolumn{2}{c}{No IDM} & \multicolumn{2}{c}{Concat IDM} & Gain \\
\cmidrule(lr){2-3}\cmidrule(lr){4-5}
& Success $\uparrow$ & NFE $\downarrow$ & Success $\uparrow$ & NFE $\downarrow$ & (pp) \\
\midrule
SIGReg & \PTPlainSuccess & \PTPlainNFE & \PTConcatSuccess & \PTConcatNFE & $+68.1$ \\
VICReg & $5.2{\pm}3.0$ & $1.039{\pm}0.059$ & $26.0{\pm}7.1$ & $0.890{\pm}0.050$ & $+20.8$ \\
\bottomrule
\end{tabular*}
\end{minipage}\hfill
\begin{minipage}[t]{0.36\linewidth}
\vspace{0pt}
\centering
\includegraphics[width=0.97\linewidth]{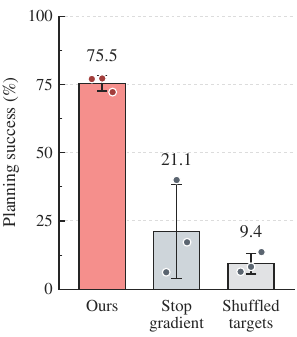}
\setlength{\abovecaptionskip}{2pt}
\captionof{figure}{\textbf{Action supervision through the encoder.} Blocking encoder gradients or shuffling action labels reduces Push-T success relative to a matched Concat IDM control ($K{=}24$).}
\label{fig:action_supervision}
\end{minipage}
\end{figure}

\paragraph{Inverse-dynamics designs.}
\label{sec:ablate}
Compared with $\Delta z$-IDM, Concat IDM has higher mean success on Push-T (\PTConcatSuccess\% versus \PTDeltaSuccess\%) and the same mean on TwoRoom (\TRConcatSuccess\% versus \TRDeltaSuccess\%).
Against PLDM, Concat IDM gains $24.0$ percentage points and wins all three matched Push-T training pairs (Table~\ref{tab:main_baselines}a).
This cross-design comparison retains PLDM's own optimization configuration.

\paragraph{Dependence on regularization.}
The benefit of inverse supervision also appears with VICReg~\citep{bardes2022vicreg}.
In a control that replaces SIGReg with VICReg and keeps the backbone, data, training schedule, and planner fixed, adding Concat IDM increases success from $5.2\%$ to $26.0\%$ and reduces NFE from $1.039$ to $0.890$ (Table~\ref{tab:regularizer_2x2}b).

\paragraph{Action supervision through the encoder.}
We isolate the contribution of the inverse loss by blocking its encoder gradients or disrupting action--transition correspondence (Figure~\ref{fig:action_supervision}).
Detaching the IDM inputs reduces success from the matched Concat IDM control of $75.5{\pm}2.8\%$ to $21.1{\pm}17.2\%$; shuffling action targets reduces it to $9.4{\pm}3.7\%$.
Both interventions also worsen NFE.
These reductions support the role of correctly paired action supervision in shaping representations useful for planning.

\subsection{Role of appearance diversity}
\label{sec:diversity}

\paragraph{A larger gain at higher visual diversity.}
\label{sec:kcurve}
On Push-T, we render the same physical trajectory pool under $K{=}1$ and $K{=}24$, matching training blocks across conditions.
The high-minus-low difference in the Concat-minus-Plain gain is \PTSuccessInteraction\ percentage points, positive in all three blocks (Figure~\ref{fig:motivation}b).

\begin{figure}[!b]
\centering
\begin{minipage}[b]{0.49\linewidth}
\centering
\includegraphics[width=\linewidth]{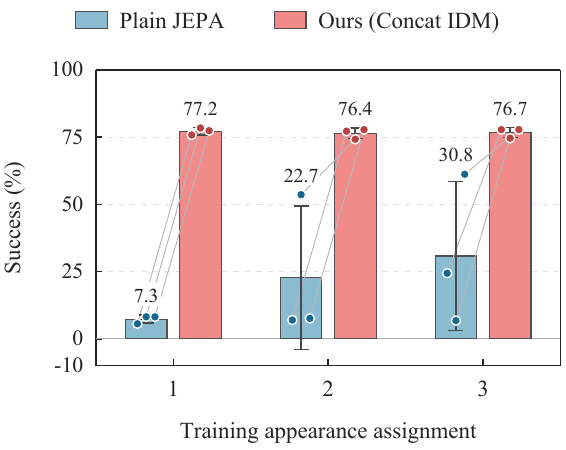}
\par\small (a) Planning success
\end{minipage}\hfill
\begin{minipage}[b]{0.49\linewidth}
\centering
\includegraphics[width=\linewidth]{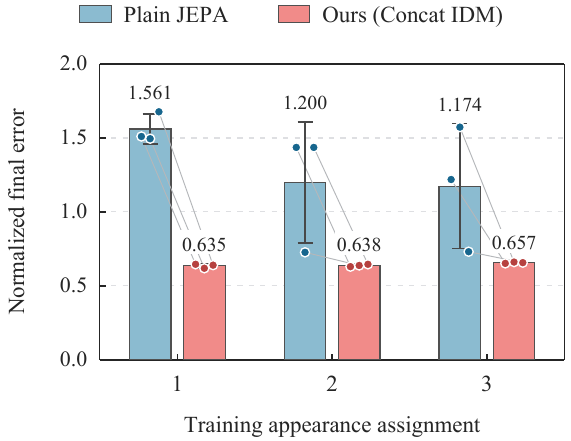}
\par\small (b) Final-state error
\end{minipage}
\caption{\textbf{Robustness to training appearance assignments on Push-T ($K{=}24$).} (a) Planning success. (b) Normalized final error. Bars show means with sample SDs; dots denote trained models and lines connect matched runs.}
\label{fig:diversity_interaction}
\end{figure}

\paragraph{Robustness across appearance assignments.}
\label{sec:assignment}
Concat IDM maintains high success when the mapping from physical trajectories to appearances changes.
We train three Plain/Concat pairs under each of three independent episode-to-appearance assignments, keeping the trajectory pool, $K{=}24$ palette, and canonical evaluation tasks and renderings fixed.
All nine Concat IDM models achieve \PTAssignConcatRange\% success, close to the canonical \PTConcatSuccess\%, while Plain JEPA spans \PTAssignPlainRange\%.
Concat IDM improves success and reduces NFE in all nine matched pairs (Figure~\ref{fig:diversity_interaction}a,b).

Changing the assignment tests a different source of variation from changing $K$.
The palette size and physical transitions remain fixed, but the grouping of trajectories that share an appearance changes.
Keeping the evaluation rendering fixed then tests whether the learned planner depends on a particular training association between appearance and behavior.
The narrow Concat IDM success range across these assignments contrasts with the broad Plain JEPA range, suggesting that action supervision reduces sensitivity to this association within the observed palette.

\subsection{Action information and planning costs}
\label{sec:mechanism}

\begin{figure}[!t]
\centering
\begin{minipage}[b]{0.49\linewidth}
\vspace{0pt}
\centering
\includegraphics[width=\linewidth]{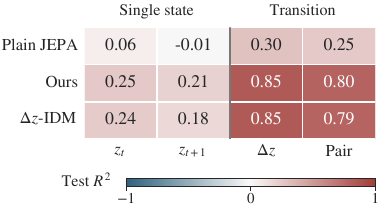}
\par\small (a) Frozen-encoder action probes
\end{minipage}\hfill
\begin{minipage}[b]{0.49\linewidth}
\vspace{0pt}
\centering
\includegraphics[width=\linewidth]{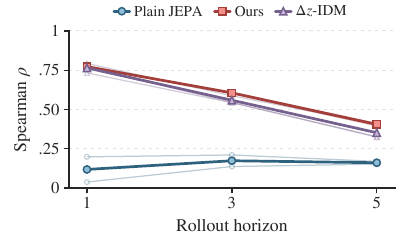}
\par\small (b) Latent-cost agreement
\end{minipage}
\caption{\textbf{Action information and latent rollout consistency on Push-T.} (a) Frozen-encoder action-probe test $R^2$; \emph{Pair} denotes $[z_t;z_{t+1}]$. (b) Spearman correlation between predicted and realized latent costs. Thin lines show individual models; thick lines show means.}
\label{fig:mechanism}
\end{figure}

\paragraph{Action information in latent transitions.}
We train fresh MLP probes on frozen encoders to predict actions from $z_t$, $z_{t+1}$, $\Delta z$, or $[z_t;z_{t+1}]$, using a common trajectory-level training, validation, and test split within the world-model data pool.
For concatenated endpoints, test $R^2$ is \PTConcatProbePair\ with Concat IDM versus \PTPlainProbePair\ with Plain JEPA (Figure~\ref{fig:mechanism}a).
Concat IDM supports displacement decoding at \PTConcatProbeDelta, compared with \PTConcatSingleProbe\ from its better individual endpoint.
The $\Delta z$-IDM encoder attains a similar displacement score (\PTDeltaProbeDelta), despite lower mean Push-T planning success (Table~\ref{tab:main_baselines}a).
The probes thus establish improved action accessibility in both representations, motivating a separate assessment of candidate-cost ordering.

\paragraph{Ordering candidate plans.}
We compare each predicted terminal cost with the realized latent cost obtained by executing the same candidate and encoding its outcome with the same model.
This diagnostic measures whether imagined and realized outcomes induce consistent candidate rankings in the learned latent space.
At horizon $H{=}5$, Concat IDM achieves Spearman correlations of $0.395$--$0.411$, compared with $0.153$--$0.170$ for Plain JEPA (Figure~\ref{fig:mechanism}b).
Together, the probes and ranking comparisons show more accessible action information and more consistent latent rollouts.

Rank correlation summarizes the entire candidate set, whereas CEM acts on its low-cost subset.
We therefore also examine selection regret and overlap between the predicted and realized top-$10\%$ candidates.
At $H{=}5$, Concat IDM has normalized selection regret of $0.961$--$1.067$, compared with $1.664$--$3.107$ for Plain JEPA; top-$10\%$ overlap increases from $0.147$--$0.170$ to $0.263$--$0.277$.
Both measures support the ranking improvement at the part of the candidate set used for action selection.
Because realized costs use each model's own encoder, these comparisons assess rollout consistency within a representation; closed-loop success separately evaluates whether optimizing that representation achieves the physical task.
The similar displacement-probe scores of the two IDM variants further show why action decodability alone cannot determine their planning performance.

\subsection{Sensitivity and scope}
\label{sec:scope}

\paragraph{IDM weight and training progress.}
At the default $\lambda{=}1$, Concat IDM improves success from $9.1\%$ to $77.2\%$ and reduces NFE from $1.353$ to $0.652$ (Table~\ref{tab:lambda_sensitivity}a).
The additional weight settings yield success rates of $71.0\%$ at $\lambda{=}0.1$ and $85.8\%$ at $\lambda{=}10$.
Across the three training runs, the advantage emerges early: at epoch~3, Concat IDM reaches $63.1\%$ success versus $10.6\%$ for Plain JEPA, with lower NFE at every reported epoch (Table~\ref{tab:training_progress_main}b).

\begin{table}[htbp]
\caption{IDM weight and training progress on Push-T ($K{=}24$). In (a), $n{=}3$ for $\lambda\in\{0,1\}$ and $n{=}1$ otherwise. In (b), each epoch uses the same three training runs per method. Success is in percent; NFE is lower-is-better.}
\label{tab:lambda_sensitivity}
\label{tab:training_progress_main}
\centering
\footnotesize
\begin{minipage}[t]{0.35\linewidth}
\centering
\vspace{0pt}
\textbf{(a) IDM loss weight}\par\smallskip
\setlength{\tabcolsep}{3pt}
\begin{tabular}{@{}crr@{}}
\toprule
$\lambda$ & Success $\uparrow$ & NFE $\downarrow$ \\
\midrule
$0$ & $9.1{\pm}2.7$ & $1.353{\pm}0.108$ \\
$0.1$ & $71.0$ & $0.713$ \\
$1$ & $77.2{\pm}0.2$ & $0.652{\pm}0.032$ \\
$10$ & $85.8$ & $0.610$ \\
\bottomrule
\end{tabular}
\end{minipage}\hfill
\begin{minipage}[t]{0.63\linewidth}
\centering
\vspace{0pt}
\textbf{(b) Training progress}\par\smallskip
\setlength{\tabcolsep}{2pt}
\begin{tabular}{@{}crrrr@{}}
\toprule
& \multicolumn{2}{c}{Plain JEPA} & \multicolumn{2}{c}{Concat IDM} \\
\cmidrule(lr){2-3}\cmidrule(lr){4-5}
Epoch & Success $\uparrow$ & NFE $\downarrow$ & Success $\uparrow$ & NFE $\downarrow$ \\
\midrule
$1$ & $4.3{\pm}1.4$ & $1.616{\pm}0.041$ & $10.6{\pm}1.6$ & $0.914{\pm}0.090$ \\
$3$ & $10.6{\pm}0.9$ & $1.287{\pm}0.088$ & $63.1{\pm}4.0$ & $0.694{\pm}0.028$ \\
$5$ & $24.4{\pm}11.4$ & $1.039{\pm}0.225$ & $70.8{\pm}1.6$ & $0.678{\pm}0.015$ \\
$10$ & $9.1{\pm}2.7$ & $1.353{\pm}0.108$ & $77.2{\pm}0.2$ & $0.652{\pm}0.032$ \\
\bottomrule
\end{tabular}
\end{minipage}
\end{table}

The training curves distinguish early gains from sustained improvement.
Plain JEPA's mean success peaks at epoch~5 among the reported checkpoints and then falls, while Concat IDM continues to improve through epoch~10.
Over the same interval, their NFE values move in opposite directions.
Action supervision therefore changes the trajectory of planning performance during training, not only the final checkpoint comparison.
The lower NFE already visible at epoch~1 also indicates progress toward the goal before either model achieves high success.

\paragraph{Scope and limitations.}
The principal finding concerns planning with seen appearances and matching current and goal renderings.
Appearance remains decodable, and success on a disjoint palette is $4.2\%$ for Plain JEPA and $13.4\%$ for Concat IDM, leaving substantial room for transfer to unseen appearances.
Directly measuring action-outcome merging in trained models would further connect the proposition to planning behavior.

\section{Conclusion}
\label{sec:concl}

Action prediction penalizes the merging of distinguishable action outcomes that forward latent prediction can leave unpenalized.
Under controlled appearance variation over fixed physical trajectories, Concat IDM improves planning, with larger gains at high visual diversity on Push-T and TwoRoom and consistent benefits across independent appearance assignments.
Encoder interventions, action probes, and candidate-cost comparisons link these gains to correctly paired action supervision, accessible transition information, and improved latent rollout consistency.
These findings motivate supervising action--consequence distinctions alongside forward prediction when learning planning representations from visually diverse observations.

\clearpage
\subsection*{AI use statement}
In this work, we used generative AI tools for drafting and editing prose, assisting with the development and checking of the theoretical analysis, and supporting code development and experiment management.
We have reviewed all AI-assisted text and code; all scientific claims, experimental protocols, numbers, and analyses were verified by the authors.
We take responsibility for the final content of this work, including text, claims, and artifacts produced with the aid of generative AI.

\subsection*{Reproducibility statement}
Code, configuration files, and evaluation scripts are planned for public release.
We report closed-loop success together with normalized final error. Continuous-error metrics use episodes with finite recorded errors and initial distance at least $10^{-6}$; success includes all episodes.
Every task, method, and appearance-set size in Table~\ref{tab:main_multirun} uses three independently trained models, each evaluated on 500 episodes.
The additional Push-T assignment study uses three training repetitions per method under each of three episode assignments on the same 500 tasks as the canonical models.
Unless otherwise specified, sensitivity analyses, diversity sweeps, and additional controls use the same training and evaluation protocol.

\bibliography{references}
\bibliographystyle{iclr2027_conference}

\end{document}